\pdfoutput=1
\documentclass{ecai} 

\usepackage{latexsym}
\usepackage{amssymb}
\usepackage{amsmath}
\usepackage{amsthm}
\usepackage{booktabs}
\usepackage{graphicx}
\usepackage{color}

\usepackage{algorithm}
\usepackage{algorithmic}
\usepackage{multirow} 
\usepackage{color}
\usepackage{enumitem}
\usepackage{url}

\newcommand{\BibTeX}{B\kern-.05em{\sc i\kern-.025em b}\kern-.08em\TeX}

\begin{document}


\begin{frontmatter}


\paperid{734} 


\title{Generalizing Visual Question Answering from Synthetic to Human-Written Questions via a Chain of QA with a Large Language Model}


\author[A]{\fnms{Taehee}~\snm{Kim}}
\author[B]{\fnms{Yeongjae}~\snm{Cho}}
\author[A]{\fnms{Heejun}~\snm{Shin}}
\author[B]{\fnms{Yohan}~\snm{Jo}}
\author[A]{\fnms{Dongmyung}~\snm{Shin}\thanks{Corresponding Author. Email: shinsae11@radisentech.com.}}

\address[A]{Radisen Co. Ltd.}
\address[B]{Seoul National University}


\begin{abstract}
Visual question answering (VQA) is a task where an image is given, and a series of questions are asked about the image. To build an efficient VQA algorithm, a large amount of QA data is required which is very expensive. Generating synthetic QA pairs based on templates is a practical way to obtain data. However, VQA models trained on those data do not perform well on complex, human-written questions. To address this issue, we propose a new method called {\it chain of QA for human-written questions} (CoQAH). CoQAH utilizes a sequence of QA interactions between a large language model and a VQA model trained on synthetic data to reason and derive logical answers for human-written questions. We tested the effectiveness of CoQAH on two types of human-written VQA datasets for 3D-rendered and chest X-ray images and found that it achieved state-of-the-art accuracy in both types of data. Notably, CoQAH outperformed general vision-language models, VQA models, and medical foundation models with no finetuning. The source code for CoQAH is available at \url{https://github.com/tae2hee/CoQAH}.
\end{abstract}

\end{frontmatter}


\section{Introduction}

Visual question answering (VQA) aims to build an automated algorithm to answer a series of questions regarding a given image. This task has a wide range of potential applications, such as interpreting medical images \cite{lin2023medical} and supporting visually impaired people \cite{bigham2010vizwiz}.

Recently, general vision language models (VLMs), trained using a massive amount of image and text data, have shown promising results in solving VQA tasks \cite{alayrac2022flamingo,achiam2023gpt,liu2024visual}. However, their performance is limited when tested in VQA tasks of specific domains (e.g., VLM model trained on large amounts of medical images and captions on the web vs. VQA model specialized for chest X-rays) \cite{li2023comprehensive}.

\begin{figure}[!t]
\centering
\includegraphics[scale=.65]{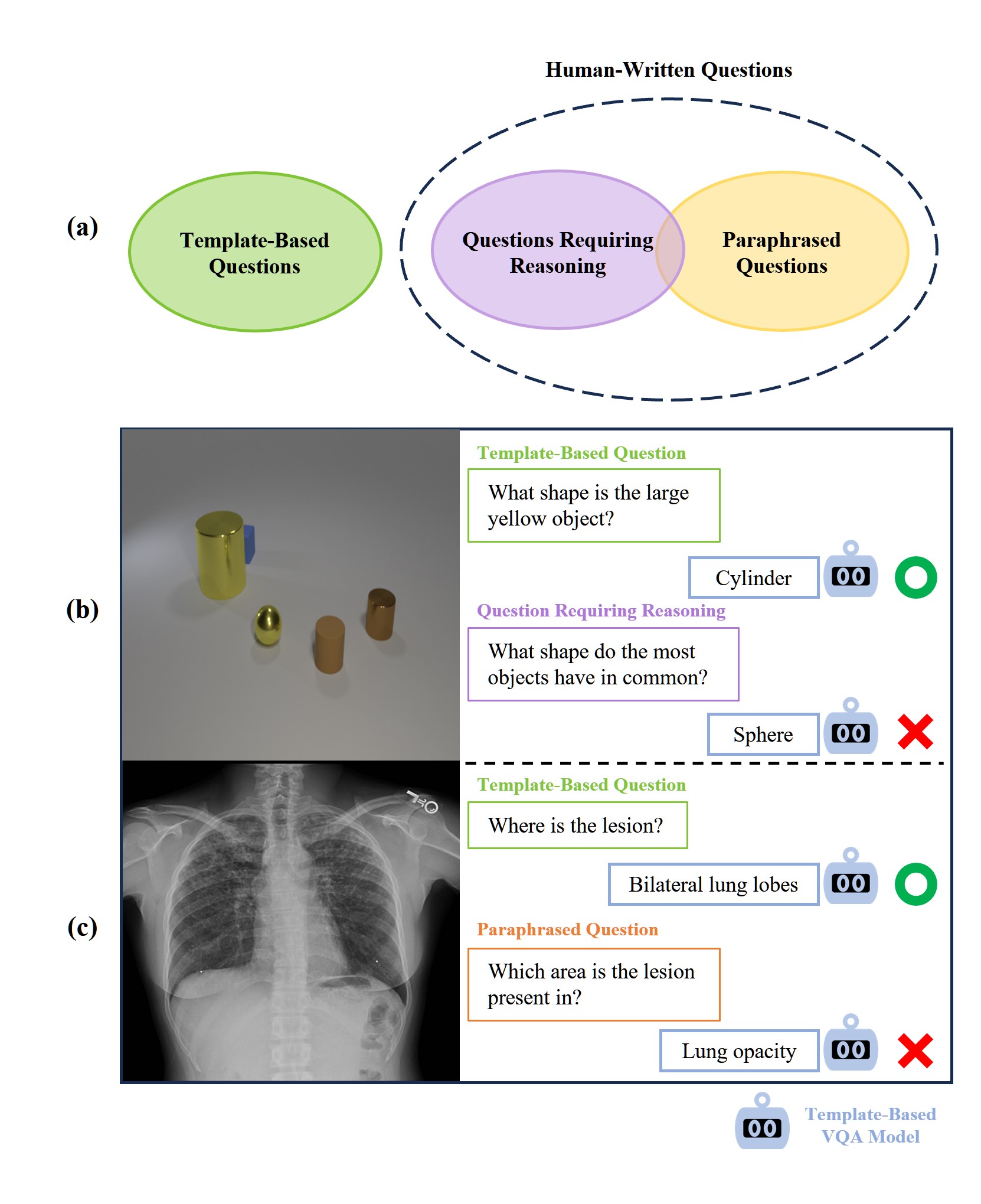}
\caption{(a) Human-written questions, compared to the fixed format of the template-based questions, include more complex free-form questions, such as ones requiring reasoning. (b), (c) Example cases where a template-based VQA model fails to answer human-written questions correctly.}
\label{fig:1}
\end{figure}

This challenge highlights the need to train or finetune VLM models using VQA data in a particular domain. However, acquiring such data is challenging, requiring experts to design VQA datasets carefully without contradiction and logical errors. For instance, to create a VQA dataset of chest X-rays, radiologists must write QA pairs consistent with the given radiographs \cite{lau2018dataset, liu2021slake}.

To address this challenge, a template-based approach, which automatically synthesizes QA pairs based on pre-defined templates, has been adopted. This approach produced high-quality VQA datasets with minimum errors (e.g., CLEVR \cite{johnson2017clevr} and MIMIC-Diff-VQA \cite{hu2023expert}). Based on these datasets, some studies reported very high accuracy on internal testing (e.g., 99.7\% in CLEVR) \cite{hu2023expert,hudson2018compositional,perez2018film,kamath2021mdetr}. However, they also found that those VQA models failed to answer human-written questions that deviated from the templates, such as questions requiring complex reasoning (e.g., \textit{what shape is the large yellow object?} vs. \textit{what shape do the most objects have in common?} in Figure~\ref{fig:1}b) and, even paraphrased questions (e.g., \textit{where is the lesion?} vs. \textit{which area is the lesion present in?} in Figure~\ref{fig:1}c).

In this study, we propose a novel method called \textit{chain of QA for human-written questions} (CoQAH) which can correctly answer complex, human-written questions beyond a fixed set of synthetic, template-based questions without the need for finetuning. In the core part of CoQAH, a large language model (LLM) sequentially asks template-based questions to a VQA model trained with synthetic data. This step-by-step QA process enables the LLM to collect valuable and accurate information about an image and reach a reasonable conclusion for a given question.

To check the effectiveness of CoQAH, we utilized two different types of template-based VQA datasets (CLEVR for 3D-rendered images and MIMIC-Diff-VQA for chest X-rays) for AI training. We then used multiple human-written VQA datasets (CLEVR-Human \cite{johnson2017inferring}, VQA-RAD \cite{lau2018dataset}, and SLAKE \cite{liu2021slake}) for AI testing. As a result, CoQAH achieved state-of-the-art performance on all the human-written VQA datasets, surpassing general VLMs and other VQA models.

Our contributions are summarized as follows: (1) we propose CoQAH, a new method that answers human-written questions through a step-by-step QA process between an LLM and a VQA model. (2) our method achieved state-of-the-art performance on multiple human-written VQA datasets, surpassing general VLMs and other VQA models. (3) we demonstrated the effectiveness of CoQAH on two types of VQA datasets, including 3D-rendered images (CLEVR and CLEVR-Human) and chest X-rays (MIMIC-Diff-VQA, VQA-RAD, and SLAKE).

\section{Related Works}

\subsection{General Vision Language Models}

A general VLM is an AI model that can understand images and texts, along with their relationships, in various tasks. One of the earliest approaches to creating a general VLM was to combine separate vision and text encoders, aligning features from both encoders \cite{radford2021learning,yu2022coca}. Another approach utilized a pre-trained LLM, which can understand the meaning and context of texts. \citet{tsimpoukelli2021multimodal} proposed training and combining a vision encoder with a frozen LLM. Similarly, \citet{alayrac2022flamingo} proposed finetuning the connecting layers between an LLM and a vision encoder. More recently, \citet{liu2024visual} proposed training VLMs using the instructions of various tasks to make a general-purpose assistant. These general VLMs can be applied to various vision-language tasks, but they often lack expertise in specific domains. CoQAH addresses this issue by employing a specialized VQA model (e.g., chest X-ray VQA) that interacts with an LLM.

\subsection{Template-Based VQA}

Creating VQA datasets manually is a time-consuming and expensive process. To overcome this challenge, some researchers have attempted to synthesize QA pairs using pre-defined templates \cite{hu2023expert,johnson2017clevr}. For instance, one study introduced a VQA dataset for chest X-rays by extracting abnormal findings from radiologist's reports \cite{hu2023expert}. Similarly, another study created 3D-rendered images of various objects and generated complex template-based questions related to the visual features of those images \cite{johnson2017clevr}. Several other studies have used such datasets to train and evaluate the visual understanding capabilities of AI models \cite{hudson2018compositional,perez2018film,kamath2021mdetr,hu2023expert}.

\subsection{LLM-Aided VQA}
Several studies have used an LLM to solve VQA problems \cite{yang2022empirical,tiong2022plug,guo2023images,zhou2023vicor,you2023idealgpt}. This is done by feeding the relevant information to the LLM, which then uses its high contextual understanding and reasoning ability to answer the question. For example, some researchers \cite{yang2022empirical,tiong2022plug,guo2023images} have proposed feeding an image description from a captioning model to an LLM, while others have utilized VQA and captioning models together to give visual clues to an LLM  \cite{zhou2023vicor}. Recently, some researchers have suggested using iterative interactions between an LLM and a VQA model to answer complex visual questions \cite{you2023idealgpt}. CoQAH uses a template-based VQA model that is specialized to questions in specific domains, which makes it different from other methods above.

\section{Method}

\begin{figure*}[h]
\centering
\includegraphics[width=\textwidth]{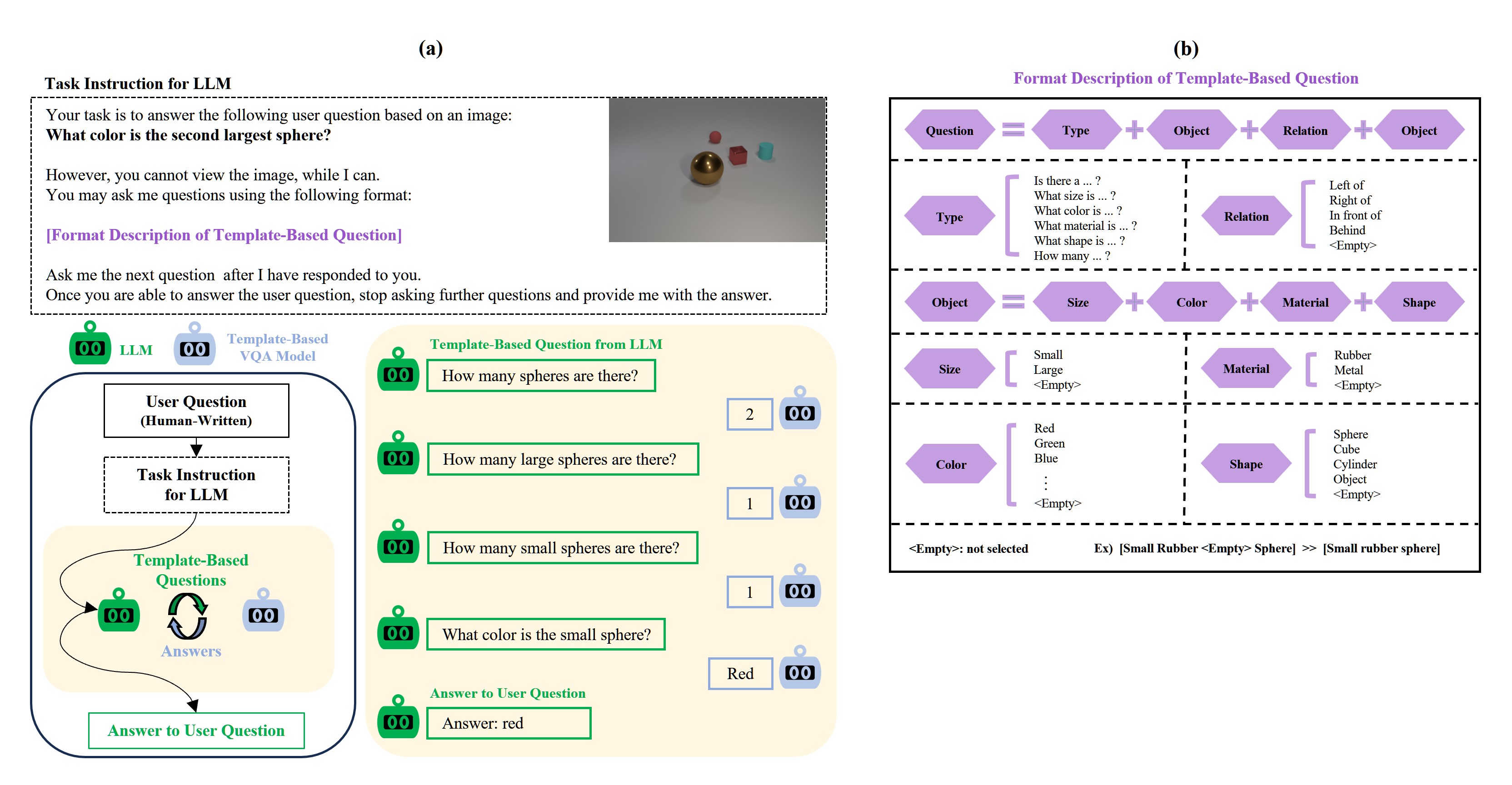}
\caption{An overview of the proposed CoQAH method. (a) An example of a task instruction is shown at the top. The figure on the left describes an overall interaction process between an LLM and a template-based VQA model to reach the final answer to the user question. On the right, the figure represents an example of dialogue between the two models. (b) The template format for the questions in the CLEVR dataset is described, which is composed of several different entities ($\langle question \rangle = \langle type \rangle + \langle object \rangle + \langle relation \rangle + \langle object \rangle$). For each entity, a few options are available to be selected (e.g., small or large or $\langle Empty \rangle$ for $\langle Size \rangle$ entity).}
\label{fig:2}
\end{figure*}

\subsection{CoQAH}
The CoQAH method aims to conclude a reasonable answer for a given complex, human-written question (i.e., user question; see Figure~\ref{fig:2}a) by employing synthetic, template-based VQA data only. To achieve this, it utilizes interactions between two sub-models, a template-based VQA model ($VQA_t$; see Supplementary Table 1 and 2 for the performance of VQA models) and a large language model ($LLM$).

Suppose a user question and an image from an human-written VQA dataset ($q_h \in Q_h,i_h \in I_h; Q_h$ for human-written questions and $I_h$ for images) are given; first, the $LLM$ is prompted with a task instruction ($\rho$; see Section \ref{sec3.2} for detail) that directs the model to generate a template-based question that conforms to the question templates used to train the VQA model ($q_t^1=LLM(\rho);$ $q_t^1 \in Q_t$ where $Q_t$ includes all possible template-based questions; see Section \ref{sec3.2}). Next, $VQA_t$ answers the question ($a_t^1=VQA_t(q_t^1,i_h)$). After that, all the previous dialogue, including the task instruction, template-based questions, and answers from $VQA_t$ $(\rho + q_t^1+a_t^1)$, is again fed into $LLM$, generating a next template-based question ($q_t^2=LLM(\rho,q_t^1,a_t^1)$). This QA process is repeated until $LLM$ has sufficient information to reach the final answer for the user question or it reaches the maximum number of questions to be queried (twenty questions for CLEVR-Human and five questions for VQA-RAD and SLAKE; see Section \ref{sec5.4}). Note that $LLM$ cannot access the image, whereas $VQA_t$ does.

\subsection{Task Instruction} \label{sec3.2}
The task instruction $\rho$ is a detailed description of how $LLM$ should respond to and interact with $VQA_t$ (Figure~\ref{fig:2}a). The instruction starts with \textit{Your task is to answer the following user question based on an image:} followed by the user question. Then, we included a statement to declare that $LLM$ cannot see a given image (\textit{However, you cannot view the image, while I can.}). After that, we asked $LLM$ to give a question that can only be generated based on a template format (\textit{You may ask me questions using the following format}; see Section \ref{sec3.3} for details). In the last part, we added some sentences to ask $LLM$ to query the next question step-by-step after having the response from $VQA_t$ (\textit{Ask me the next question after I have responded to you.}) and conclude the answer for the user question whenever it is ready (\textit{Once you are able to answer the user question, stop asking further questions and provide me with the answer.}).

\subsection{Template-Based Questions from LLM} \label{sec3.3}
Figure~\ref{fig:2}b describes how we instructed $LLM$ to generate a template-based question for the CLEVR dataset \cite{johnson2017clevr}. We first let $LLM$ know the structure of the question, which is composed of several different entities ($\langle question \rangle = \langle type \rangle + \langle object \rangle + \langle relation \rangle + \langle object \rangle$ in Figure~\ref{fig:2}b). Subsequently, we asked $LLM$ to choose one of the options available for each entity (e.g., small or large or $\langle Empty \rangle$ for $\langle Size \rangle$ entity). The template format for MIMIC-Diff-VQA is shown in Supplementary Figure 1. The detailed description of the task instructions for both CLEVR and MIMIC-Diff-VQA is displayed in Supplementary Figure 2 and 3.

\subsection{Existence and Uniqueness Handler}

\begin{figure*}[h]
\centering
\includegraphics[width=1\textwidth]{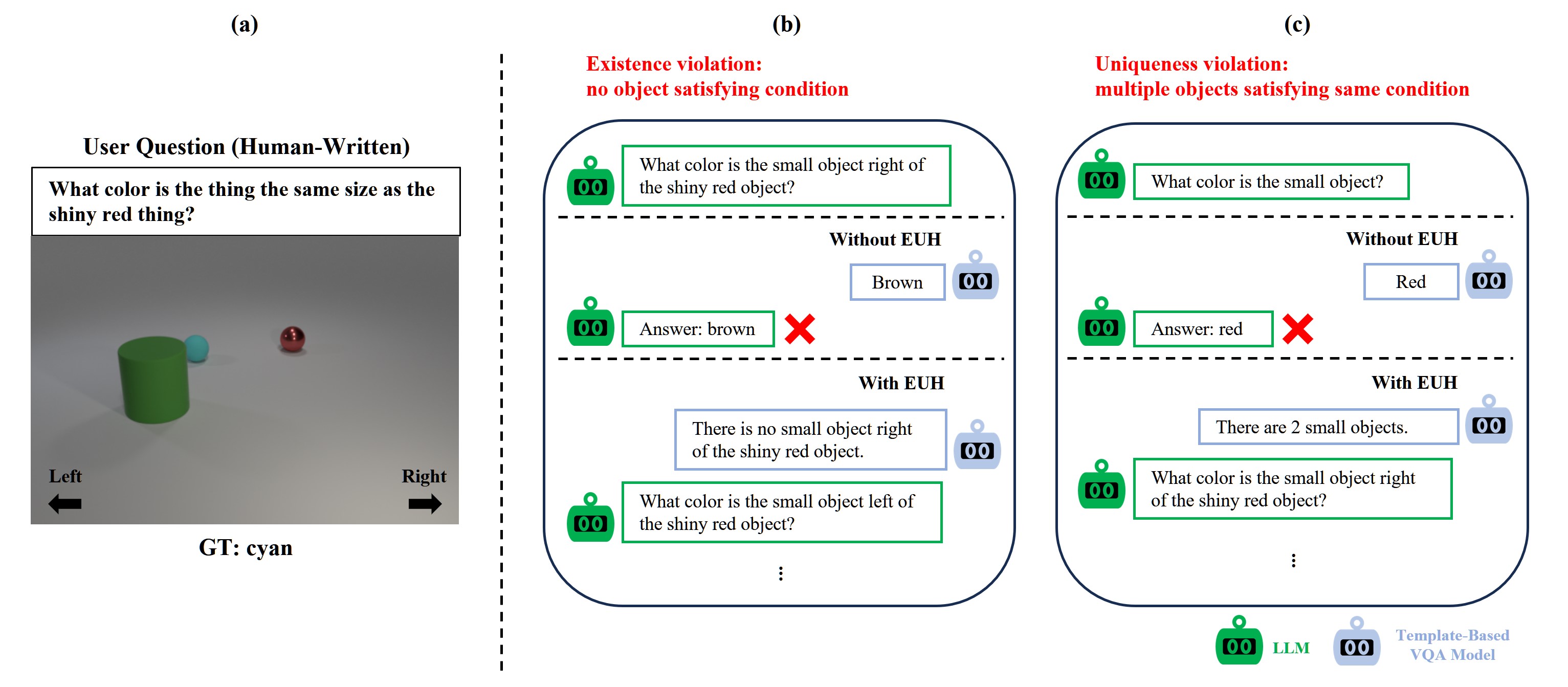}
\caption{Illustration of how the existence and uniqueness handler (EUH) can prevent an LLM from concluding an incorrect answer for a given user question. (a) An example case includes a question, an answer, and an image. (b) A dialogue between an LLM and a VQA model when the existence of an object is violated. With EUH, the VQA model checks the presence of an object and lets LLM know its existence. (c) A dialogue when the uniqueness of an object is violated. With EUH, the VQA model successfully lets LLM know the number of objects satisfying the same condition.}
\label{fig:3}
\end{figure*}

Even if $LLM$ successfully generated a template-based question, in some cases, we observed that there were still some questions that had logical errors (see Figure~\ref{fig:3}): questions that ask the property of an object that does not exist in the image (existence violation in Figure~\ref{fig:3}b) and ask the property of multiple objects with the same condition (uniqueness violation in Figure~\ref{fig:3}c). For these types of questions, $VQA_t$ provided incorrect answers unexpectedly.

To address this problem, we introduced an existence and uniqueness handler (EUH), which continually checks the violation of both conditions on a template-based question generated from $LLM$. As described in Algorithm~\ref{alg:algorithm}, we first extracted the object entity ($\langle object \rangle + \langle relation \rangle + \langle object \rangle$) from the question ($\langle type \rangle + \langle object \rangle + \langle relation \rangle + \langle object \rangle$). For example, the object entity ("small object" + "right of" + "shiny red object") was extracted from the question (\textit{What color is the small object right of the shiny red object?}). Then, we checked the existence of the object entity by querying to $VQA_t$ (e.g., \textit{Is there a small object right of the shiny red object?}). When the existence of the object was confirmed, we continued to check the uniqueness by asking $VQA_t$ to count the number of objects (e.g., \textit{How many small objects right of the shiny red object are there?}). Using EUH, we prevented $LLM$ from directly providing a wrong answer without considering logical contradiction.

\raggedbottom



\begin{algorithm}[h]
    \caption{Existence and uniqueness handler}
    \label{alg:algorithm}
    \textbf{Input}: Question from LLM $Q_L$; Input image $I$; VQA model $VQA(\cdot)$; A function that extracts $\langle \textit{Object} \rangle + \langle \textit{Relation} \rangle + \langle \textit{Object} \rangle$ part from the question $ \langle \textit{Type} \rangle + \langle \textit{Object} \rangle+ \langle \textit{Relation} \rangle+ \langle \textit{Object} \rangle$ $Extract\_Entity(\cdot)$\\
    \textbf{Output}: Answer to $Q_L$ $A_L$
    \begin{algorithmic}[1] 
        \STATE $E \gets Extract\_Entity(Q_L)$
        \STATE $Q_E \gets $ ``Is there a" $+$ $E$ $+$ ``?"
        \STATE $A_E \gets VQA(Q_E, I)$
        \IF{$A_E$ is ``No"}
            \STATE $A_L \gets $ ``There is no" $+$ $E$
        \ELSIF{$A_E$ is ``Yes"}
            \STATE $Q_U \gets $ ``How many" $+$ $E$ $+$ ``are there?"
            \STATE $A_U \gets VQA(Q_U, I)$
            \IF{$A_U$ is not ``1"}
                \STATE $A_L \gets $ ``There are" $+$ $A_U$ $+$ $E$
            \ELSIF{$A_U$ is ``1"}
                \STATE $A_L \gets VQA(Q_L, I)$
            \ENDIF
        \ENDIF
        \STATE \textbf{Return} $A_L$
    \end{algorithmic}
\end{algorithm}


\section{Experiment Setup}

\subsection{Datasets}
For datasets, we collected two types of images (and question-answer pairs), including 3D-rendered images and chest X-ray images, to show the effectiveness of CoQAH in different image types. We first trained VQA models using template-based datasets (CLEVR \cite{johnson2017clevr} for 3D-rendered and MIMIC-Diff-VQA \cite{hu2023expert} for chest X-ray) and externally validated the model's performance on the human-written datasets (CLEVR-Human \cite{johnson2017inferring} for 3D-rendered; VQA-RAD \cite{lau2018dataset} and SLAKE \cite{liu2021slake} for chest X-ray).

\paragraph{CLEVR} is a synthetic VQA dataset (70,000 image and QA pairs for training, 15,000 for validation, and 15,000 for testing). 3D objects of different sizes, colors, materials, and shapes were rendered for each image. Then, questions were generated based on a template-based algorithm, and the answers for those questions were automatically selected among a set of candidates (e.g., \textit{3}, \textit{blue}, \textit{cylinder}, etc.). 

\paragraph{CLEVR-Human} is a variant of the CLEVR dataset that replaced questions and answers in some images of CLEVR using human questioners (17,817 for training; 7,202 for validation). The questioners created new human-written questions without any format restriction, while the answers for the questions were chosen among the same candidates in CLEVR. Currently, a test set is not available. Therefore, we reported AI performance using the validation set.

\paragraph{MIMIC-Diff-VQA} is a chest X-ray VQA dataset. QA pairs were generated using an automatic template algorithm based on radiologists' reports (700,703 QA pairs from 164,324 reports and images; 80\% for training, 10\% for validation, and 10\% for testing).

\paragraph{VQA-RAD and SLAKE} are medical image VQA datasets where clinical trainees (for VQA-RAD) or physicians (for SLAKE) wrote human-written questions and answers manually for each image. We used only the chest X-rays in those datasets (107 images for VQA-RAD and 179 images for SLAKE) for AI evaluation. In both datasets, answers are categorized as either closed-form or open-form: the answers of the closed-form questions were required to be one between two or three options (e.g., yes or no; 511 questions for VQA-RAD and 663 questions for SLAKE), whereas the answers of the open-form questions did not have any formatting restrictions (283 questions for VQA-RAD and 1,459 questions for SLAKE).

\subsection{Metrics}
We measured exact-match accuracy for the human-written questions that select answers from predefined candidates (all questions of CLEVR-Human and closed-form questions of VQA-RAD and SLAKE). For the open-form questions of VQA-RAD and SLAKE, we calculated LAVE$_{GPT-4}$ \cite{manas2024improving} that uses an LLM (GPT-4 in this paper) for the evaluation of answers.


\begin{table*}[h]
\caption{Comparison of accuracy between the general VLMs, template-based VQA models, and finetuned VQA models. CoQAH achieved the highest accuracy with large gaps compared to all the other general VLMs and template-based VQA models, although it did not utilize any data from CLEVR-Human for training. * indicates accuracy measured using the CLEVR-Human validation data.}
\centering
\begin{tabular}{lccc}
\toprule

& Trained on  & Finetuned on & Acc. on \\
& CLEVR? & CLEVR-Human? & CLEVR-Human (\%) \\

\midrule
\textbf{\textit{General VLMs}} & & & \\
LLaVA (7B) \cite{liu2024visual} & No & No & \;\;43.8$^*$ \\
LLaVA (13B) \cite{liu2024visual} & No & No & \;\;47.6$^*$ \\
GPT-4-vision \cite{achiam2023gpt} & No & No & \;\;60.1$^*$ \\

\midrule
\textbf{\textit{Template-Based VQA Models}} & & & \\
FiLM \cite{perez2018film} & Yes & No & 56.6 \\ 
MAC \cite{hudson2018compositional} & Yes & No & 57.4 \\ 
MDETR \cite{kamath2021mdetr} & Yes & No & 59.9 (59.5$^*$) \\ 
\textbf{CoQAH} & Yes & No & \;\;\textbf{74.3}$^*$ \\ 

\midrule
\textcolor[gray]{0.5}{\textbf{\textit{Finetuned VQA Models}}} & & & \\
\textcolor[gray]{0.5}{FiLM \cite{perez2018film}} & \textcolor[gray]{0.5}{Yes} & \textcolor[gray]{0.5}{Yes} & \textcolor[gray]{0.5}{75.9} \\ 
\textcolor[gray]{0.5}{MAC \cite{hudson2018compositional}} & \textcolor[gray]{0.5}{Yes} & \textcolor[gray]{0.5}{Yes} & \textcolor[gray]{0.5}{81.5} \\ 
\textcolor[gray]{0.5}{MDETR \cite{kamath2021mdetr}} & \textcolor[gray]{0.5}{Yes} & \textcolor[gray]{0.5}{Yes} & \textcolor[gray]{0.5}{81.7} \\ 

\bottomrule
\end{tabular}
\label{tab1}
\end{table*}



\begin{table*}[h]
\caption{Comparison of the performance between the medical foundation and template-based VQA models on VQA-RAD and SLAKE. CoQAH reported the highest accuracy and LAVE$_{GPT-4}$ in both open- and closed-form questions. There were huge gaps between the LAVE$_{GPT-4}$ of the OFA-MIMIC and CoQAH, meaning that the chain of QA process in CoQAH indeed improved the correctness of the answers significantly.}
\centering
\begin{tabular}{lcccc}
\toprule

& \multicolumn{2}{c}{VQA-RAD} & \multicolumn{2}{c}{SLAKE} \\
\cmidrule{2-3} \cmidrule{4-5}
& Acc. on & LAVE$_{GPT-4}$ & Acc. on & LAVE$_{GPT-4}$ \\
& Closed-Form (\%) & on Open-Form & Closed-Form (\%) & on Open-Form \\

\midrule

\textbf{\textit{Medical Foundation Models}} & & & & \\
BiomedGPT \cite{zhang2023biomedgpt} & 43.2 & 0.150 & - & - \\ 
Med-Flamingo \cite{moor2023med} & 46.4 & 0.184 & 33.2 & 0.140 \\ 
MedVInT-TD \cite{zhang2023pmc} & 57.3 & 0.274 & 46.5 & 0.396 \\ 

\midrule
\textbf{\textit{Template-Based VQA Models}} & & & & \\
OFA-MIMIC \cite{wang2022ofa} & 59.5 & 0.095 & 69.4 & 0.097 \\ 
\textbf{CoQAH} & \textbf{67.5} & \textbf{0.302} & \textbf{73.9} & \textbf{0.425} \\ 

\bottomrule

\end{tabular}
\label{tab2}
\end{table*}


\subsection{Benchmarks}

\subsubsection{Benchmarks  for CLEVR-Human}
To benchmark the CLEVR-Human dataset, we employed three types of AI models, including general VLMs, VQA models trained with CLEVR (i.e., template-based VQA), and finetuned with CLEVR-Human (i.e., finetuned VQA). For the general VLMs, we utilized LLaVA \cite{liu2024visual} (version 1.5; 7B and 13B parameters; source code and model weights available on GitHub; see task instruction in Supplementary Figure 4) and GPT-4-Vision \cite{achiam2023gpt} (version 1106-preview; API used; see task instruction in Supplementary Figure 5), testing them on the CLEVR-Human validation dataset. For the template-based VQA models, we benchmarked FiLM \cite{perez2018film}, MAC \cite{hudson2018compositional}, MDETR \cite{kamath2021mdetr}, and our CoQAH. In CoQAH, we combined MDETR trained with CLEVR as a VQA model and GPT-4 as an LLM, measuring accuracy on the CLEVR-Human validation data. Finally, we compared the reported scores from the previous studies for the accuracy of the finetuned VQA models (FiLM, MAC, and MDETR).

\subsubsection{Benchmarks for VQA-RAD and SLAKE}
To benchmark the VQA-RAD and SLAKE datasets, we employed three general VLMs specialized for the medical domain (i.e., medical foundation models): BiomedGPT \cite{zhang2023biomedgpt} (no task instruction needed), Med-Flamingo \cite{moor2023med} (see task instruction in Supplementary Figure 6), and MedVInT-TD \cite{zhang2023pmc} (see task instruction in Supplementary Figure 7). As template-based VQA models, we benchmarked OFA \cite{wang2022ofa} trained by ourselves (i.e., OFA-MIMIC; using a model weight of OFA$_{base}$ with 182M parameters; AdamW optimizer with learning rate = 1e-4, $\beta_1$ = 0.9, and $\beta_2$ = 0.999; batch size = 16) because, so far, no VQA model trained with MIMIC-Diff-VQA is available to report performance on VQA-RAD and SLAKE. In CoQAH, we combined the OFA-MIMIC and GPT-4. Note that, different from the benchmarks on CLEVR-Human, we could not finetune the OFA-MIMIC using VQA-RAD or SLAKE because the number of chest X-rays is very limited (107 X-rays for VQA-RAD and 179 X-rays for SLAKE). 

\subsubsection{Post-Processing of Answers}
We observed that some of the answers the AI models provided frequently had the same meaning but different expressions (e.g., \textit{radiograph} vs. \textit{x-ray}, \textit{pa views} vs. \textit{pa}, \textit{right side} vs. \textit{right}, etc.), resulting in misleading scores. Therefore, we substituted those answers as standard forms using post-processing (see Supplementary Table 3). We applied this post-processing to all the AI models we tested for fair comparison.

\section{Results}

\subsection{Benchmarks on CLEVR-Human}
Table~\ref{tab1} compares accuracy between the general VLMs, template-based, and finetuned VQA models. Overall, the general VLMs reported lower accuracy than the template-based VQA models, except for the latest GPT-4-vision (e.g., 60.1\% for GPT-4-vision vs. 59.9\% for MDETR). CoQAH achieved the highest accuracy with large gaps compared to all the general VLMs and template-based VQA models (e.g., 74.3\% accuracy for CoQAH vs. 59.9\% for MDETR). There are still some performance gaps between CoQAH and the finetuned models (e.g., 74.3\% for CoQAH vs. 81.7\% for finetuned MDTER). However, it is worth noting that obtaining a large number of human-written QA pairs and finetuning a model is very difficult for most VQA tasks in general, and our framework substantially improves VQA performance using only synthetic QA pairs that are much easier to collect.

\begin{figure*}[h]
\centering
\includegraphics[width=\textwidth]{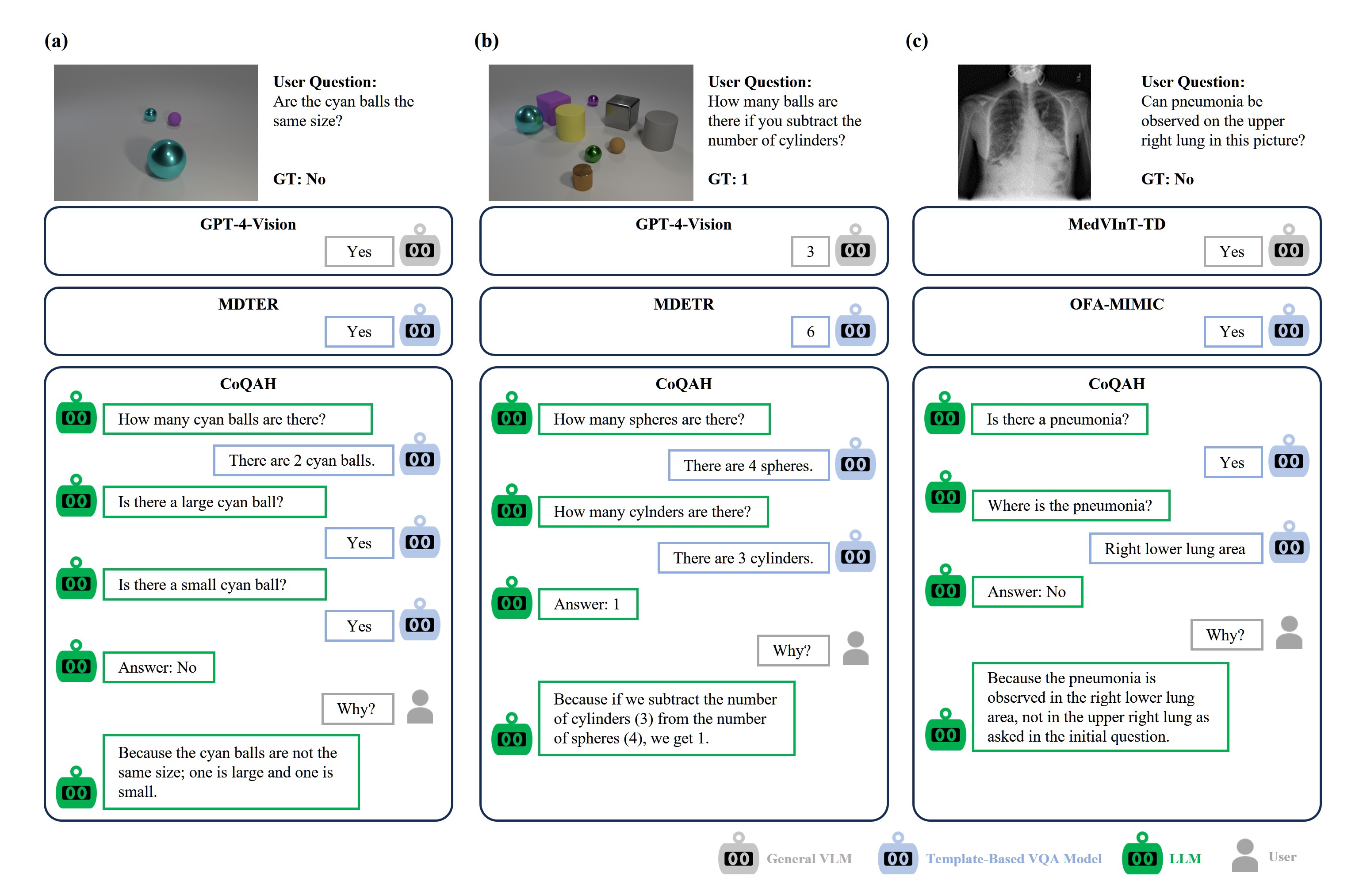}
\caption{Example dialogues of CoQAH when we asked an LLM to interpret the reasons for the answers. In all cases, CoQAH successfully delineated the rationale behind the answers based on the information collected through the dialogues. For instance, for the question of \textit{Are the cyan balls are the same size?}, CoQAH correctly explained the reason (\textit{one is large and one is small}) for the response (\textit{No}) based on the information from the two previous questions (\textit{Is there a large cyan ball?} and \textit{Is there a small cyan ball?}).}
\label{fig:4}
\end{figure*}

\subsection{Benchmarks on VQA-RAD and SLAKE}
Table~\ref{tab2} summarizes the performance of the medical foundation and template-based VQA models for VQA-RAD and SLAKE. Among all the models, CoQAH reported the highest accuracy in the closed-form questions (e.g., VQA-RAD: 67.5\% for CoQAH vs. 59.5\% for OFA-MIMIC, SLAKE: 73.9\% for CoQAH vs. 69.4\% for OFA-MIMIC), and also the highest LAVE$_{GPT-4}$ in the open-form questions (e.g., VQA-RAD: 0.302 for CoQAH vs. 0.274 for MedVInT-TD, SLAKE: 0.425 for CoQAH vs. 0.396 for MedVInT-TD). There were huge gaps between the LAVE$_{GPT-4}$ of the OFA-MIMIC and CoQAH, meaning that the chain of QA process in CoQAH indeed improved the correctness of the answers significantly.

\subsection{Interpreting Reasons for Answers}
Since the CoQAH method utilizes an LLM to understand the context and finally answer the user question, we can ask the rationale of the final answer to the LLM (i.e., interpreting the AI model's behavior), simply querying `\textit{why?}'. For a few complex questions, we asked the LLM to explain the reasons to check whether it had logically driven their conclusions.

As shown in the examples of Figure~\ref{fig:4}, CoQAH successfully delineated the rationale behind the answers based on the information collected through the dialogues. For example, for the question of whether there are two cyan balls of the same size (Figure~\ref{fig:4}a), CoQAH reasonably explained the reason (\textit{one is large and one is small}) for the answer (\textit{No}) after checking the size of the cyan balls (\textit{Is there a large cyan ball?} and \textit{Is there a small cyan ball?}). More examples can be found in Supplementary Figure 8 (for 3D-rendered images) and 9 (for chest X-rays).

\subsection{Effect of the Maximum Number of Questions} \label{sec5.4}
The maximum number of questions to be queried is important parameter since it can directly affect the quality of the answers from CoQAH. Therefore, we measured the accuracy of CoQAH on the CLEVR-Human validation dataset (5\% used), changing that number from 0 to 30. As shown in Table~\ref{tab3}, The maximum accuracy was achieved when the maximum number of questions was 20, and even if we increased the number to 30, the accuracy was the same.

\subsection{Ablation Study}
We conducted an ablation study by excluding several components in CoQAH and measured the accuracy based on the CLEVR-Human validation data (5\% used). The results of the ablation study are summarized in Table~\ref{tab4}. When we ablated EUH in CoQAH, the accuracy on CLEVR-Human was slightly decreased (75.8\% for CoQAH vs. 74.7\% without EUH). When ablating the step-by-step QA process in CoQAH by enforcing the LLM to ask all the questions simultaneously (\textit{ask me up to 20 questions all at once}), the accuracy was degraded mainly (75.8\% for CoQAH vs. 63.1\% without the process), indicating the importance of this process to derive the answers correctly. Finally, when we ablated the format description in the task instruction by letting LLM generate any form of questions during the chain of QA process, the accuracy was dramatically degraded (75.8\% for CoQAH vs. 35.8\% without the format description). This means that the questions from the LLM must conform to the template format so the VQA model can understand those questions. Additionally, to investigate the effect of few-shot prompting on CoQAH, we compared the performance of CoQAH in zero-shot, one-shot, and two-shot settings. Dialogues between the VQA model and the LLM were provided as few-shot exemplars to the LLM. The results are shown in Table ~\ref{tab5}. The performance increased when few-shot exemplars were given, but increasing the number of exemplars from one to two did not further improve performance.   

\begin{table}[t]
\caption{Accuracy on CLEVR-Human validation dataset according to the changes of the maximum number of questions to be queried in CoQAH. We only used 5\% of the validation data for the ablation study. Until the maximum number of questions reached 20, accuracy increased as the number of questions increased. However, the accuracy remained the same when the maximum number of questions was set at 20 and 30.}
\centering
\begin{tabular}{cccccc}
\toprule

Maximum number of  & 
\multirow{2}{*}{0} &
\multirow{2}{*}{5} &
\multirow{2}{*}{10} &
\multirow{2}{*}{20} &
\multirow{2}{*}{30} \\
questions to be queried \\ 

\midrule

Acc.on & 
\multirow{2}{*}{60.0} &
\multirow{2}{*}{68.1} &
\multirow{2}{*}{69.4} &
\multirow{2}{*}{75.8} &
\multirow{2}{*}{75.8} \\
CLEVR-Human (\%) \\

\bottomrule
\end{tabular}
\label{tab3}
\end{table}



\begin{table}[t]
\caption{Comparison of accuracy on CLEVR-Human by ablating the step-by-step QA process, EUH, or the template format description. Ablating the QA process degraded the accuracy significantly. Ablating the format description  dramatically degraded the accuracy, meaning that the questions from the LLM must conform to the template format.}
\centering
\begin{tabular}{lc}
\toprule

& Acc. on \\
& CLEVR-Human (\%) \\

\midrule


CoQAH & 75.8 \\
without EUH & 74.7 \\
without step-by-step QA process & 63.1 \\
without format description & 35.6 \\

\bottomrule
\end{tabular}
\label{tab4}
\end{table}


\begin{table}[t]
\caption{The performance of CoQAH in zero-shot, one-shot, and two-shot settings. Dialogues between the VQA model and the LLM were provided as few-shot exemplars to the LLM. Performance increased when few-shot exemplars were provided, but adding more exemplars, from one to two, did not result in any further improvement.}
\centering
\begin{tabular}{cccc}
\\
\toprule

Prompt Setting  & Zero-shot & One-shot & Two-shot \\

\midrule

Acc. on & \multirow{2}{*}{75.8} & \multirow{2}{*}{80.0} & \multirow{2}{*}{79.4} \\
CLEVR-Human (\%) \\

\bottomrule
\end{tabular}
\label{tab5}
\end{table}


\subsection{Error Analysis of CoQAH}
We performed an error analysis by investigating the dialogues where CoQAH failed to answer correctly (5\% CLEVR-Human validation used) and categorized them according to the reasons (e.g., incorrect answers from VQA model). The results of the error analysis are shown in Table~\ref{tab6}.

The most significant portion of errors (48\%) occurred when the LLM failed to query key questions or made a hasty conclusion after asking some irrelevant questions (e.g., Supplementary Figure 8e). It means that the LLM still has shown a limited reasoning ability during the QA process for some complex questions. The second largest portion (25\%) was due to incorrect responses from the VQA model during the QA process (e.g., Supplementary Figure 8d). Some errors (11\%) dedicated to the questions where the LLM was impossible to answer by querying template-based questions only (e.g., Supplementary Figure 8f). Other errors (15\%) included failures of following instructions, data format (e.g., \textit{The answer should be in \{\}}), and so on.


\begin{table}[t]
\caption{Results of the error analysis for CoQAH. We investigated the dialogues where CoQAH failed to answer correctly and categorized them according to the reasons. The most significant portion of errors occurred when the LLM failed to query key questions or made a hasty conclusion after asking some irrelevant questions, showing a limited reasoning ability.}
\centering
\begin{tabular}{lc}
\toprule

Error Category & Percentage (\%) \\

\midrule

Limited reasoning ability of LLM & 48 \\
& \\
Incorrect response from VQA model & 25 \\
& \\
Impossible to answer by querying & \multirow{2}{*}{11} \\
template-based questions only & \\
& \\
Other errors & 15 \\

\bottomrule
\end{tabular}
\label{tab6}
\end{table}


\section{Conclusion}
In this study, we proposed a novel CoQAH methodology that answers human-written questions via a reasoning process based on a dialogue. Throughout the extensive experiments, we have shown that CoQAH achieved state-of-the-art performance on various datasets of different images (3D-rendered and chest X-ray) without finetuning using the human-written questions. We believe that this new method potentially promotes the wide applications of AI for visual question answering, improving the practicality of AI in real environment.

\section{Limitation}
First, we could only test a limited number of general VLM models (LLaVA \cite{liu2024visual} and GPT-4-Vision \cite{achiam2023gpt}) for CLEVR-Human since, currently, many LLMs do not support taking an image as an input. Second, we reported the accuracy of the AI models we ran on CLEVR-Human validation data because the test set was unavailable. Third, even though we tried our best to optimize the task instructions for the general VLM and medical foundation models, those might not be the best due to the lack of methods to tune the instructions. Fourth, we only validated CoQAH using two types of images (3D-rendered and chest X-ray) because of a limited number of template-based datasets.

\appendix
\bibliography{mybibfile}

\clearpage

\newcommand{\beginsupplement}{
    \renewcommand{\tablename}{Supplementary Table}
    \setcounter{table}{0}
    \renewcommand{\thetable}{\arabic{table}}
    \renewcommand{\figurename}{Supplementary Figure}
    \setcounter{figure}{0}
    \renewcommand{\thefigure}{\arabic{figure}}
}

\beginsupplement
\onecolumn

\section*{Supplementary Materials}


\begin{table*}[h]
\caption{The performance of VQA models on the CLEVR testing dataset. Regardless of the models, the accuracy values were very high ($>$ 97\%), meaning that those VQA models can precisely answer the template-based questions. We used the MDETR model as a template-based VQA model for CoQAH.}
\centering
\begin{tabular}{lc}
\\
\toprule

Model & Accuracy (\%) \\

\midrule

FiLM \cite{perez2018film} & 97.7 \\
MAC \cite{hudson2018compositional} & 98.9 \\
MDETR \cite{kamath2021mdetr} & 99.7 \\

\bottomrule
\end{tabular}
\label{sup:tab1}
\end{table*}



\begin{table*}[h]
\caption{The performance of VQA models on the MIMIC-Diff-VQA testing dataset. Since only one VQA model (EKAID), trained with MIMIC-Diff-VQA, was available, we trained an OFA model by ourselves using a model weight of OFA$_{base}$ with 182M parameters (i.e., OFA-MIMIC). Compared to the EKAID, OFA-MIMIC showed higher performance in all the metrics. We used the OFA-MIMIC as a template-based VQA model for CoQAH.}
\centering
\begin{tabular}{lccccccc}
\\
\toprule

Model  & BLEU-1 & BLEU-2 & BLEU-3 & BLEU-4 & METEOR & ROUGE & CIDEr \\

\midrule

EKAID \cite{hu2023expert} & 0.624 & 0.541 & 0.477 & 0.422 & 0.337 & 0.645 & 1.893 \\
OFA-MIMIC \cite{wang2022ofa} & 0.662 & 0.588 & 0.528 & 0.473 & 0.362 & 0.716 & 2.332 \\

\bottomrule
\end{tabular}
\label{sup:tab2}
\end{table*}


\begin{figure*}[hbt!]
\centering
\includegraphics[width=0.35\textwidth]{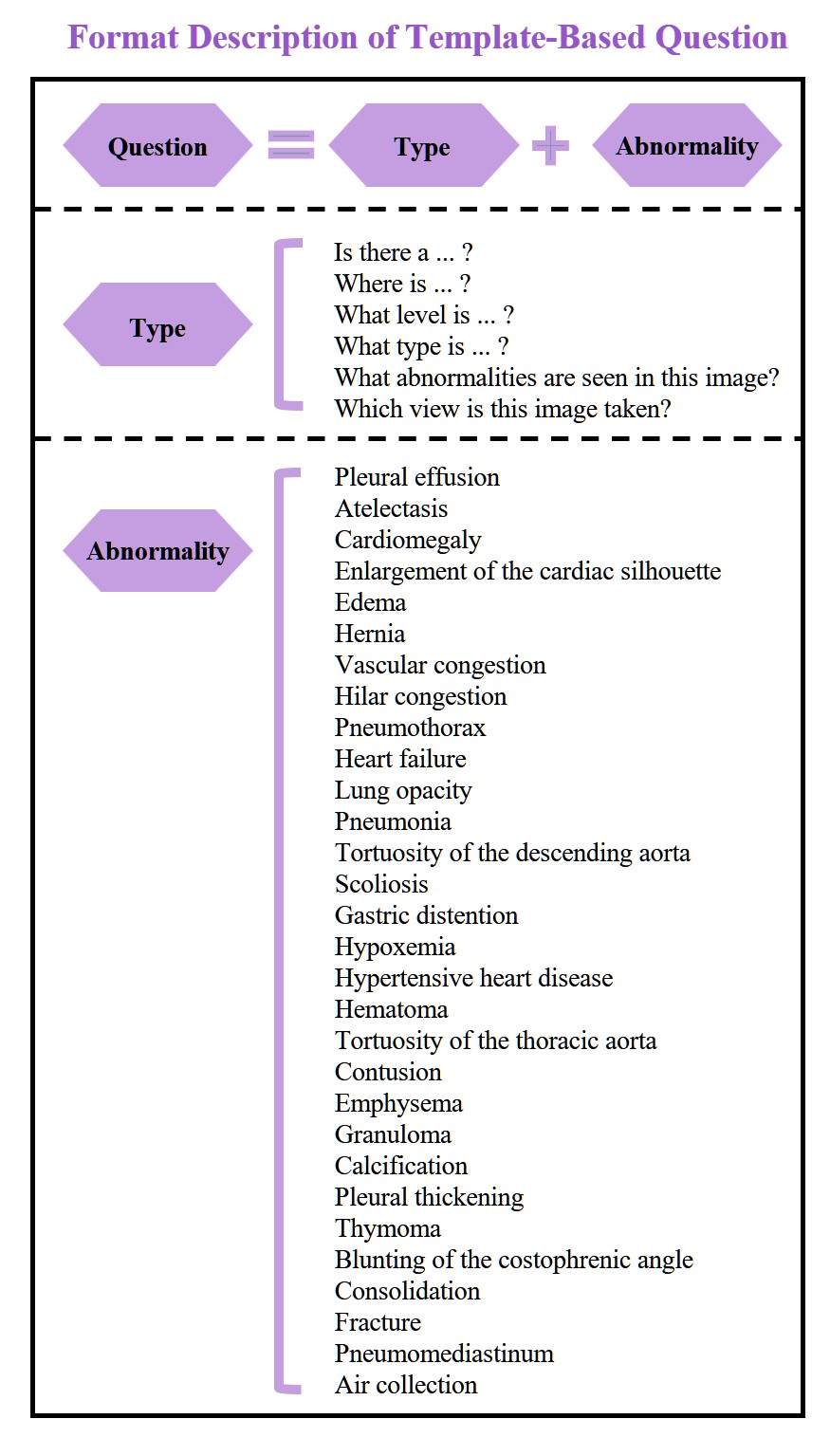}
\caption{Template format for the chest X-ray VQA dataset, MIMIC-Diff-VQA. A question is only composed of two parts, $\langle type \rangle$ and $\langle abnormality \rangle$. Similar to the case of the CLEVR dataset, we asked an LLM to select an option available for each $\langle type \rangle$ and $\langle abnormality \rangle$ to generate a template-based question.}
\label{sup:fig:1}
\end{figure*}

\begin{figure*}[h]
\centering
\includegraphics[width=\textwidth]{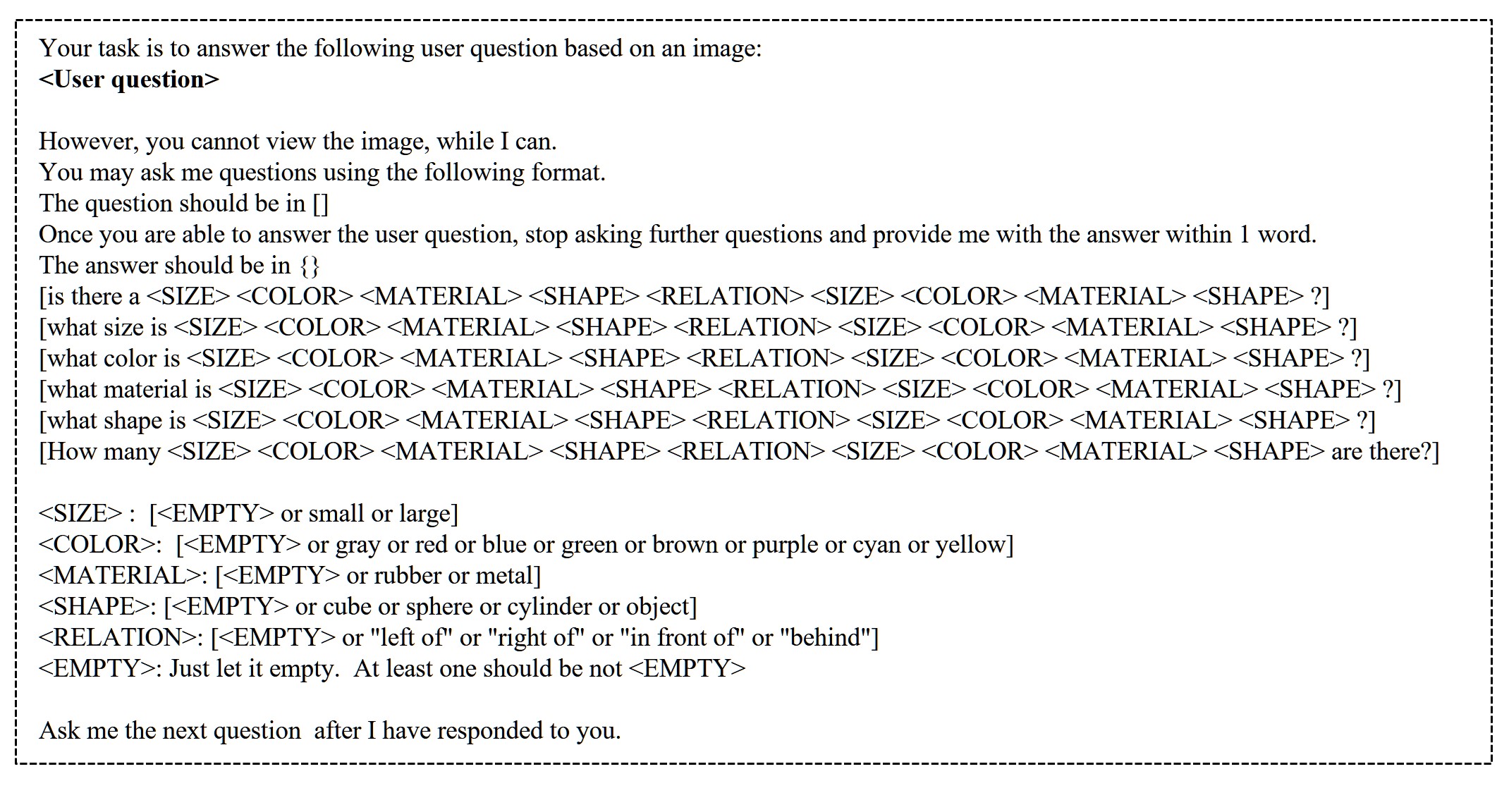}
\caption{Task instruction of CoQAH for CLEVR.}
\label{sup:fig:2}
\end{figure*}

\begin{figure*}[h]
\centering
\includegraphics[width=\textwidth]{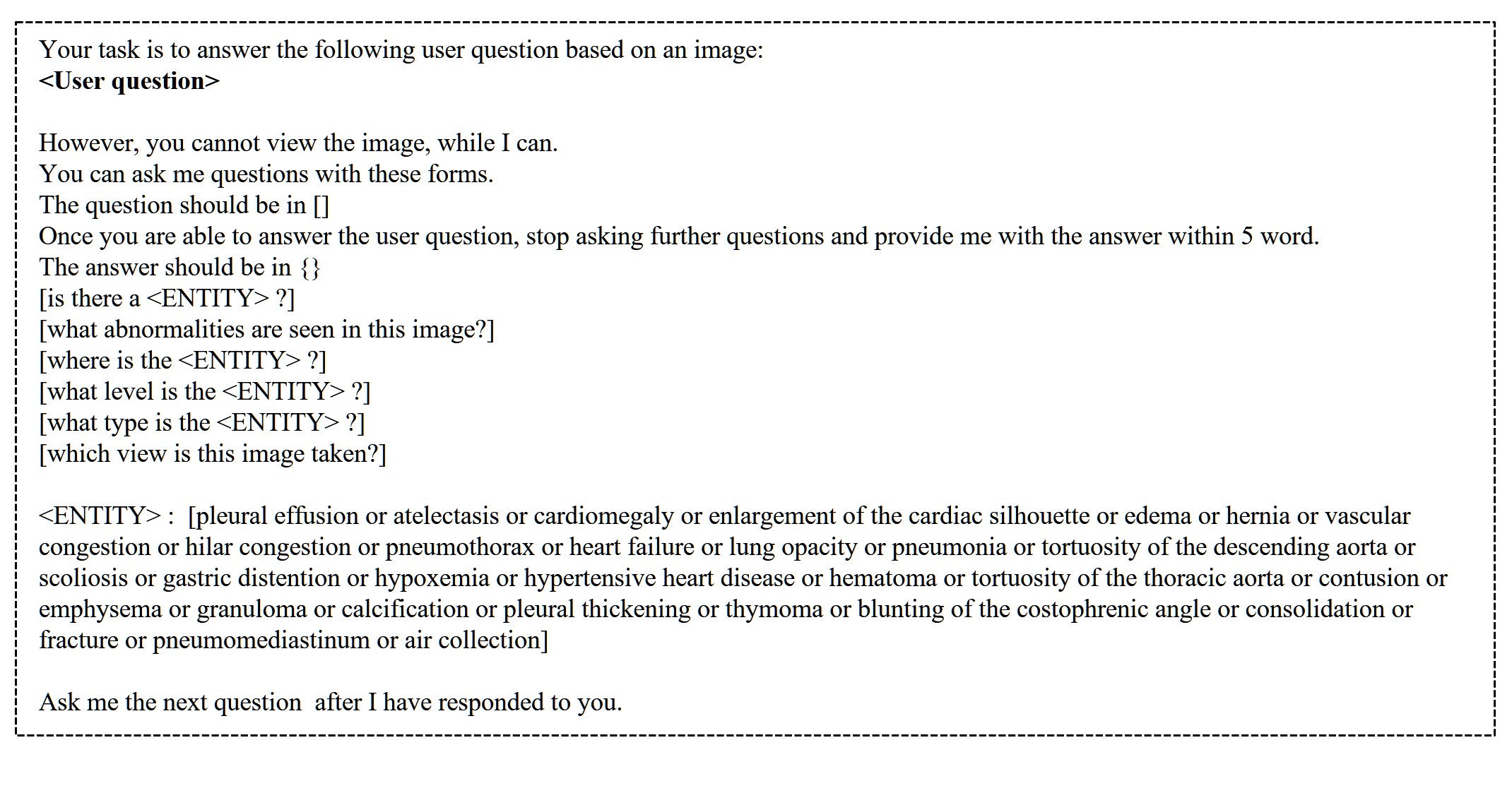}
\caption{Task instruction of CoQAH for MIMIC-Diff-VQA.}
\label{sup:fig:3}
\end{figure*}

\begin{figure*}[h]
\centering
\includegraphics[width=\textwidth]{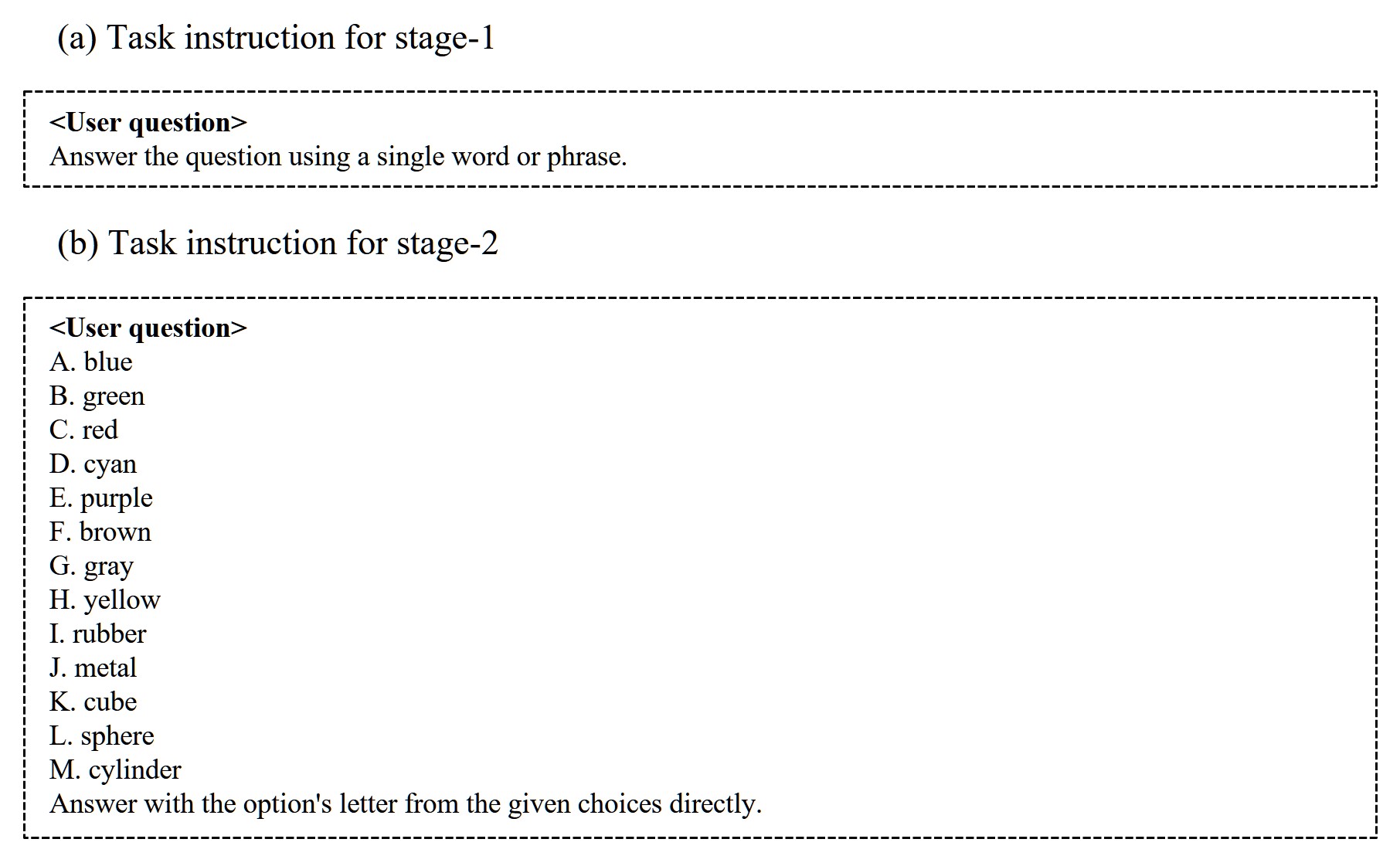}
\caption{Task instruction of LLaVA for CLEVR-Human. (a) In the first stage, we asked LLaVA to produce an answer as a single word or phrase. (b) In some cases, LLaVA provided answers of the same meaning but different expressions. Therefore, in the second stage, we corrected them by asking LLaVA to choose one of the candidates (e.g., round for the first answer vs. sphere for the corrected answer). Note that when we gave all options for the answer at once, the performance of LLaVA was primarily degraded.}
\label{sup:fig:4}
\end{figure*}

\begin{figure*}[h]
\centering
\includegraphics[width=\textwidth]{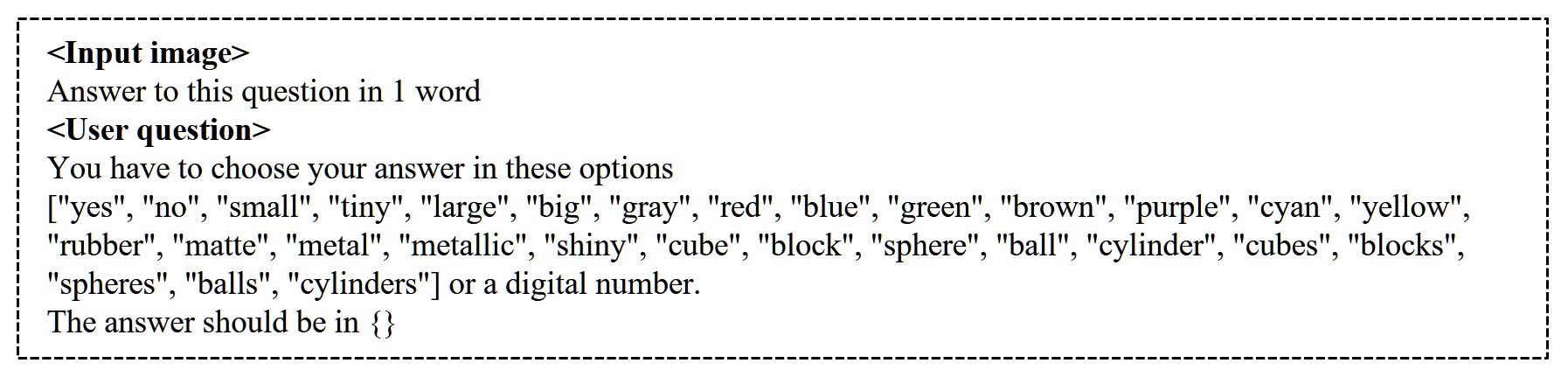}
\caption{Task instruction of GPT-4-Vision for CLEVR-Human. We provided a set of answers.}
\label{sup:fig:5}
\end{figure*}

\clearpage

\begin{figure*}[h]
\centering
\includegraphics[width=\textwidth]{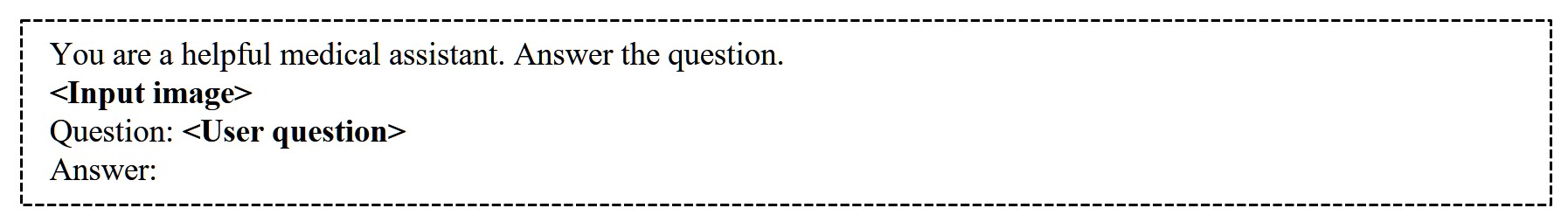}
\caption{Task instruction of Med-Flamingo for VQA-RAD and SLAKE. We limited the maximum number of output words to one for the closed-form questions and five for the open-form questions.}
\label{sup:fig:6}
\end{figure*}

\begin{figure*}[h]
\centering
\includegraphics[width=\textwidth]{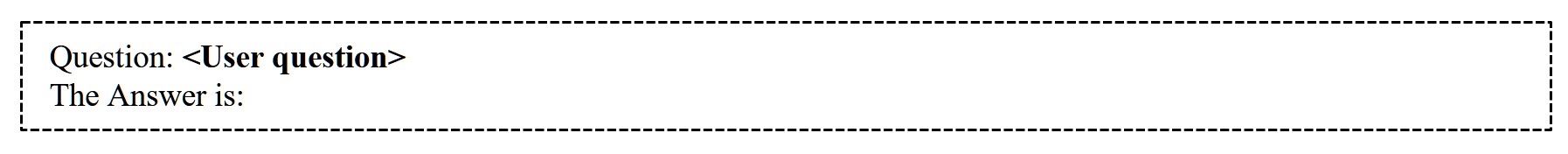}
\caption{Task instruction of MedVInT-TD for VQA-RAD and SLAKE. We limited the maximum number of output words to one for the closed-form questions and five for the open-form questions.}
\label{sup:fig:7}
\end{figure*}


\begin{table*}[hbt!]
\caption{Table summarizing the answers of AI we substituted when calculating accuracy. We changed them because those words had the same meaning but different expressions, resulting in misleading performance metrics.}
\centering
\begin{tabular}{cc}
\\
\toprule

\multicolumn{2}{c}{Synonym standardization for CLEVR-Human} \\
\midrule
Before standardization & After standardization \\
\midrule
tiny & small \\
big & large \\
matte & rubber \\
metallic & metal \\
shiny & metal \\
block & cube \\
ball & sphere \\

\bottomrule

\\
\\

\toprule

\multicolumn{2}{c}{Synonym standardization for VQA-RAD and SLAKE} \\
\midrule
Before standardization & After standardization \\
\midrule
x-ray & xray \\
x ray & xray \\
radiography & xray \\
radiograph & xray \\
cxr & chest xray \\
pa view & pa \\
ap view & ap \\
right side & right \\
left side & left \\

\bottomrule

\end{tabular}
\label{sup:tab3}
\end{table*}


\begin{figure*}[h]
\centering
\includegraphics[width=\textwidth]{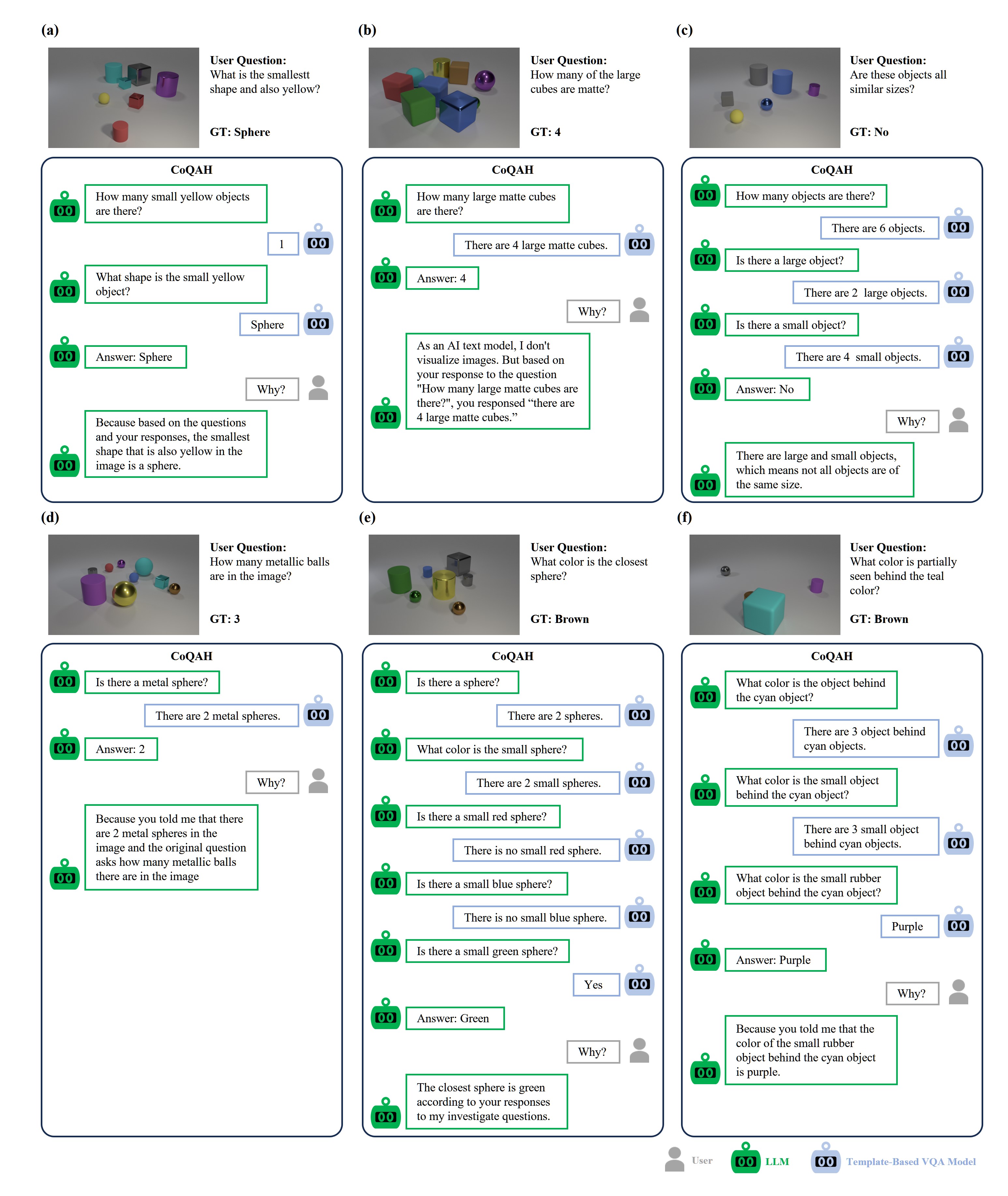}
\caption{Example dialogues of CoQAH for CLEVR-Human. (a)-(c) The cases show that CoQAH correctly answered the questions and explained the reasons. (d) The answer was wrong because the template-based VQA model failed to give correct information. (e) LLM failed to query key questions and made a hasty conclusion. (f) LLM was impossible to answer the question by querying template-based questions only.}
\label{sup:fig:8}
\end{figure*}

\begin{figure*}[h]
\centering
\includegraphics[width=\textwidth]{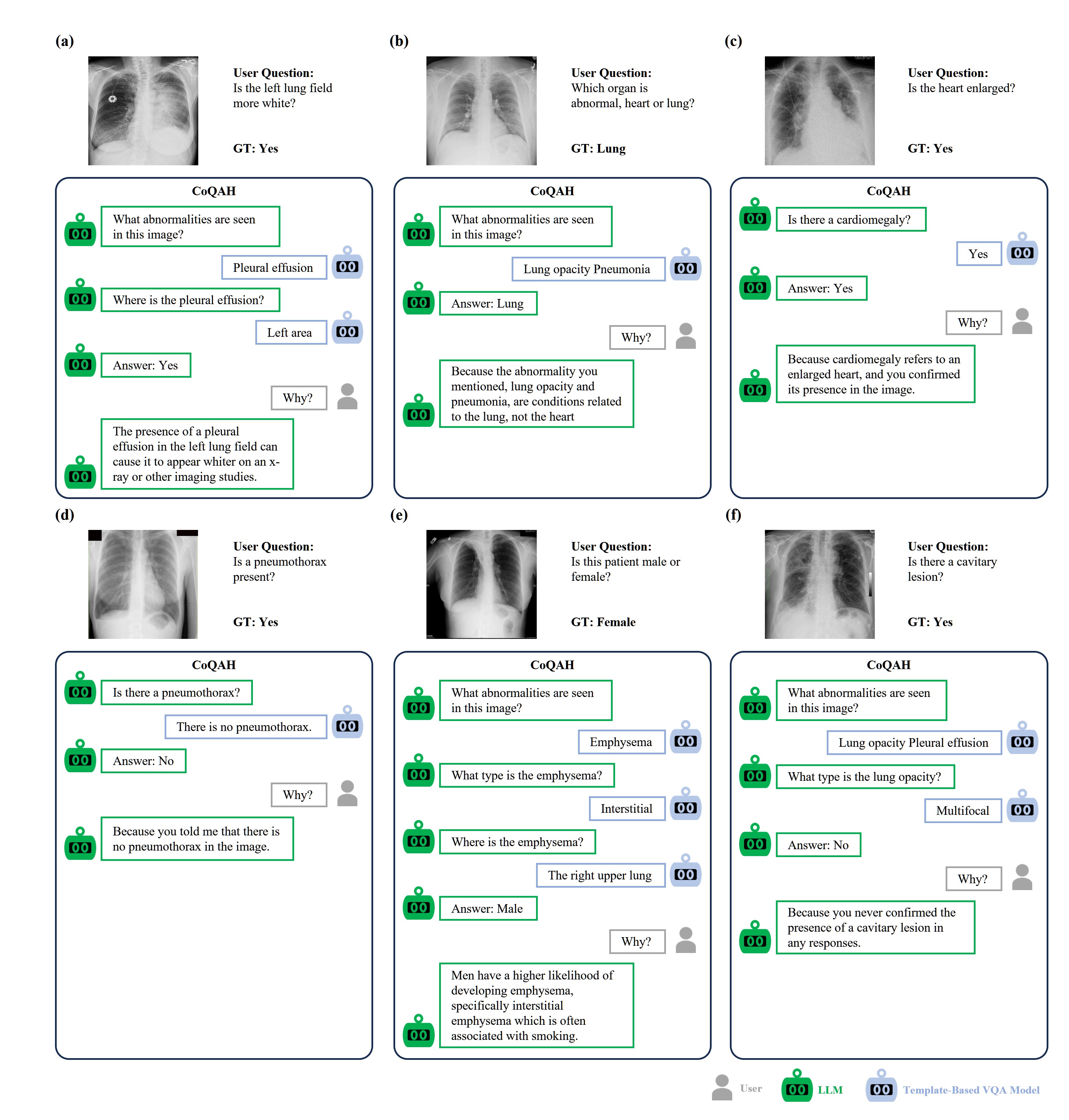}
\caption{Example dialogues of CoQAH for VQA-RAD and SLAKE. (a)-(c) CoQAH successfully answered the given user questions. (d) The VQA model failed to give the correct answer, and CoQAH failed accordingly. (e) The case shows that the large language model logically derived the answer even though the answer was incorrect. (f) Since the VQA model could not detect the lesion, CoQAH also gave the wrong answer.}
\label{sup:fig:9}
\end{figure*}

\clearpage

\end{document}